\documentclass[11pt]{article}
 
\usepackage{arxiv}

\usepackage{hyperref}       
\usepackage{url}            
\usepackage{booktabs}       
\usepackage{amsfonts}       
\usepackage{nicefrac}       
\usepackage{microtype}      
\usepackage{graphicx}
\usepackage[numbers]{natbib}
\usepackage{doi}

\usepackage{multirow}
\usepackage{algorithm}
\usepackage{algpseudocode}
\usepackage{graphicx}
\usepackage{xcolor}
\usepackage{soul}
\usepackage{amsmath}
\usepackage{amsfonts}
\usepackage{caption,subcaption}
\usepackage[para]{footmisc}
\usepackage{booktabs}
\usepackage{xcolor,colortbl}
\usepackage{hyperref}
\hypersetup{colorlinks = true, citecolor = blue }
\usepackage{cleveref}
\usepackage{caption}
\newcommand\blfootnote[1]{%
  \begingroup
  \renewcommand\thefootnote{}\footnote{#1}%
  \addtocounter{footnote}{-1}%
  \endgroup
}

\usepackage[utf8]{inputenc}
\usepackage[LFE,LAE,T1]{fontenc}
\usepackage{arabtex}
\usepackage{utf8}

\title{Unveiling Black-boxes: Explainable Deep Learning Models for Patent Classification}

\renewcommand{\shorttitle}{Explainable Deep Learning Models for Patent Classification}

\author{Md Shajalal$^{1,2}$, Sebastian Denef$^{3}$, Md. Rezaul Karim$^{4}$,  Alexander Boden$^{1,5}$, Gunnar Stevens$^{2}$\\
    $^{1}$Fraunhofer-Institute for Applied Information Technology FIT, Germany \\
    $^{2}$University of Siegen, Germany\\
    $^{3}$AGENTS.inc, Berlin, Germany\\
    $^{4}$RWTH Aachen University, Aachen, Germany\\
    $^{5}$Bonn-Rhein-Sieg University of Applied Sciences, Germany
}

\begin{document}
\maketitle

\textbf{Abstract.}
Recent technological advancements have led to a large number of patents in a diverse range of domains, making it challenging for human experts to analyze and manage. State-of-the-art methods for multi-label patent classification rely on deep neural networks~(DNNs), which are complex and often considered black-boxes due to their opaque decision-making processes. In this paper, we propose a novel deep explainable patent classification framework by introducing layer-wise relevance propagation~(LRP) to provide human-understandable explanations for predictions. We train several DNN models, including Bi-LSTM, CNN, and CNN-BiLSTM, and propagate the predictions backward from the output layer up to the input layer of the model to identify the relevance of words for individual predictions. Considering the relevance score, we then generate explanations by visualizing relevant words for the predicted patent class. Experimental results on two datasets comprising two-million patent texts demonstrate high performance in terms of various evaluation measures. The explanations generated for each prediction highlight important relevant words that align with the predicted class, making the prediction more understandable. Explainable systems have the potential to facilitate the adoption of complex AI-enabled methods for patent classification in real-world applications.\blfootnote{This is the ``\textit{Submitted Manuscript}'' to the 1$^{st}$ World Conference on eXplainable Artificial Intelligence~(xAI2023), Lisbon, Portugal. The published manuscript by Springer can be found here \url{https://doi.org/10.1007/978-3-031-44067-0_24}}


\section{Introduction}
Patent classification is an important task in the field of intellectual property management, involving the categorization of patents into different categories based on their technical contents~\cite{kucer2022deeppatent}. Traditional approaches to patent classification have relied on manual categorization by experts, which can be time-consuming and subjective~\cite{li2018deeppatent}. However, due to the exponential growth of patent applications in recent times, it has become increasingly challenging for human experts to classify patents. The international patent classification (IPC) system, which consists of 645 labels for the general classes and over 67,000 labels for the sub-groups, reflects the magnitude of challenges in multi-level patent classification tasks~\cite{kucer2022deeppatent}. Furthermore, patent texts are generally lengthy and contain irregular scientific terms, making them a challenging field of application for text classification approaches, as patents often include highly technical and scientific terms that are not commonly used in everyday language, and authors often use jargon to make their patents unique and innovative~\cite{lee2020patent}. These factors contribute to the significant challenges associated with patent classification, making it a formidable task.

However, recent advancements in machine learning~(ML) and deep neural network~(DNN) have made significant progress in automating the patent classification process. In the past, classical ML models, such as support vector machine~(SVM), K-nearest neighbour, and naive bayes, have been widely used to automatically classify patent texts~\cite{d2013text}. However, more recently, several DNN models have been proposed to address the challenges associated with patent classification. Generally, these models represent patent text using word embedding and transformer-based pre-trained models~\cite{luo2020deep,jiang2022deep,kucer2022deeppatent,li2018deeppatent,chen2020deep}. The DNN models, including recurrent neural networks~(RNN) and their variants such as convolutional neural networks~(CNN), long short-term memory networks~(LSTM), bidirectional LSTM~(Bi-LSTM), and gated recurrent unit~(GRU), can learn to classify patents based on their textual content~\cite{luo2020deep,chen2020deep,li2018deeppatent,fang2021patent2vec,haghighian2022patentnet}. Hence, these enable faster and more reliable categorization of patents and scientific articles.

Mathematically, DNN-based classification approaches are often complex in their architecture, and the decision-making procedures can be opaque~\cite{shrikumar2017learning,lundberg2017unified}. While these approaches may exhibit efficient performance in classifying patents, the decisions they make are often not understandable to patent experts, or even to practitioners of artificial intelligence~(AI). As a result, it is crucial to ensure that the methods and decision-making procedures used in patent classification are transparent and trustworthy, with clear explanations provided for the reasons behind each prediction. This is particularly important because patents are legal documents, and it is essential to comprehend the reasoning behind the classification decisions made by the model. Therefore, patent classification models should be designed to be explainable, allowing the reasons and priorities behind each prediction to be presented to users. This will help build trust in the predictive models and promote transparency among users and stakeholders.

For text-based uni-modal patent classification tasks, explanations can be provided by highlighting relevant words and their relevance to the prediction, thus increasing trust of users in the accuracy of predictions. In recent years, there has been a growing interest in developing explainable artificial intelligence (XAI) to unveil the black-box decision-making process of DNN models in diverse fields, including image processing~\cite{bach2015pixel}, text processing, finance~\cite{kute2021deep,shajalal2022explainable}, and health applications~\cite{yang2022unbox,adadi2020explainable}. These XAI models can provide insights into the decision-making process, explaining the reasoning behind specific predictions, the overall model's priorities in decision making, and thereby enhancing the transparency and trustworthiness of the application~\cite{lundberg2017unified,ribeiro2016should,shrikumar2017learning,binder2016layer,bach2015pixel}.

In this paper, our goal is to develop a patent classification framework that not only predicts the classes of patents but also provides explanations for the predicted classes. To achieve this, we propose a new explainable method for patent classification based on layer-wise relevance propagation~(LRP). This method can break down the contribution of patent terms that are crucial in classifying a given patent into a certain class. We start by representing the patent terms using a high-dimensional distributed semantic feature vector obtained from pre-trained word-embedding models. Next, we proceed to train several DNN-based models, including Bi-LSTM, CNN, and CNN-BiLSTM, which are capable of predicting the patent class. Finally, the LRP-enabled explanations interface highlights relevant words that contributed to the final prediction, providing an explanation for the model's decision.

We conducted experiments using two benchmark patent classification datasets, and the experimental results demonstrated the effectiveness of our approach in both classifying patent documents and providing explanations for the predictions. Our contributions in this paper are twofold: 
\begin{enumerate}
    \item We propose an LRP-based explainability method that generates explanations for predictions by highlighting relevant patent terms that support the predicted class.
    \item Our developed DNN models show effective performance in terms of multiple evaluation metrics on two different benchmark datasets, and performance comparison with existing works confirms their consistency and effectiveness.
\end{enumerate}

Overall, explainable DNN models offer promising solutions for patent classification, enabling faster and more accurate categorization while providing insights into the decision-making process. With the increasing volume of patent applications, the development of such explainable models could be beneficial in automatically categorizing patents with efficiency and transparency.

The rest of the paper is structured as follows: section~\ref{relatedWork} presents the summary of existing research on patent classification. Our proposed explainable deep patent classification framework is presented in section~\ref{ourApproach}. We demonstrate the effectiveness of our methods in classifying patents and explaining the predictions in detail in section~\ref{experiments}. Finally, section~\ref{conclusion} concluded our findings with some future directions in explainable patent classification research.

\section{Related Work} \label{relatedWork}
In recent years, the patent classification task has gained significant attention in the field of natural language processing (NLP) research, as evidenced by several notable studies~\cite{shalaby2018lstm,li2018deeppatent,lee2020patent}. Various methods have been employed for classifying
and analyzing patent data, and the methods can be categorized based on different factors such as the techniques utilized, the tasks' objectives~(e.g., multi-class or multi-level classification), and the type of resources used to represent the patent data~(i.e., uni-modal vs multi-
modal)~\cite{roudsari2020multi,chen2020deep,haghighian2022patentnet}. However, traditional approaches have relied on classical ML and bag-of-words~(BoW)-based text representation, which have limitations in capturing semantic and contextual information of the text, as they can only capture lexical information. With the advent of different word-embedding techniques such as \emph{word2vec} by Mikolov et al.~\cite{le2014distributed,mikolov2013distributed}, \emph{Glove} by Pennington et al.~\cite{pennington2014glove}, and \emph{FastText} by Bojanowski et al.~\cite{bojanowski2017enriching}, the NLP research has been revolutionized with the ability to represent text using high-dimensional semantic vector representations~\cite{shajalal2018sentence,shajalal2019semantic,shajalal2020coverage}. More recently, there has been a growing trend in employing transformer-based pre-trained models, including deep bidirectional transformer~(BERT)~\cite{devlin2018bert}, robust optimized BERT~(RoBERTa)~\cite{liu2019roberta}, distilled BERT~(DistilBERT)~\cite{sanh2019distilbert}, and XLNet~\cite{yang2019xlnet}, for text representation in NLP tasks.

Shaobo et al.~\cite{li2018deeppatent} introduced a deep patent classification framework that utilized convolutional neural networks~(CNNs). They started by representing the text of patents, which was extracted from the title and abstract of the USPTO-2 patent collection, using a skip-gram-based word-embedding model~\cite{li2018deeppatent}. They then used the resulting high-dimensional semantic representations to train CNN model. Similarly, Lee et al.~\cite{lee2020patent} also employed a CNN-based neural network model, however, they fine-tuned a pre-trained BERT model for text representations. A DNN-based framework employing Bi-LSTM-CRF and Bi-GRU-HAN models has been introduced to extract semantic information from patents' texts~\cite{chen2020deep}. 

A multi-level classification framework~\cite{haghighian2022patentnet} has been proposed utilizing fine-tuned transformer-based pre-trained models, such as BERT, XLNet, RoBERTa, and ELECTRA\cite{clark2020electra}. Their findings revealed that XLNet outperformed the baseline models in terms of classification accuracy. In another study, Roudsari et al.~\cite{roudsari2020multi} addressed multi-level~(sub-group level) patent classification tasks by fine-tuning a DistilBERT model for representing patent texts. Jiang et al.~\cite{jiang2022deep} presented a multi-modal technical document classification technique called \textit{TechDoc}, which incorporated NLP techniques, such as word-embedding, for extracting textual features and descriptive images to capture information for technical documents. They modelled the classification task using CNNs, RNNs, and Graph neural networks~(GNNs). Additionally, Kang et al.~\cite{kang2020prior} employed a multi-modal embedding approach for searching patent documents.

A patent classification method called \textit{Patent2vec} has been introduced, which leverages multi-view patent graph analysis to capture low-dimensional representations of patent texts~\cite{fang2021patent2vec}. Pujari et al.~\cite{pujari2021multi} proposed a transformer-based multi-task model~(TMM) for hierarchical patent classification, and their experimental results showed higher precision and recall compared to existing non-neural and neural methods. They also proposed a method to evaluate neural multi-field document representations for patent text classification. Similarly, Aroyehun et al.~\cite{aroyehun2021leveraging} introduced a hierarchical transfer and multi-task learning approach for patent classification, following a similar methodology. Roudsari et al.~\cite{roudsari2021comparison} compared different word-embedding methods for patent classification performance. Li et al.~\cite{li2022copate} proposed a contrastive learning framework called \textit{CoPatE} for patent embedding, aimed at capturing high-level semantics for very large-scale patents to be classified. An automated ensemble learning-based framework for single-level patent classification is introduced by Kamateri et al.~\cite{kamateri2022automated} .

However, to the best of our knowledge, none of the existing patent classification methods are explainable. Given the complexity of the multi-level classification task, it is crucial for users and patent experts to understand the reasoning behind the AI-enabled method's predictions, as it classifies patents into one of more than 67,000 classes~(including sub-group classes). Therefore, the aim of this paper is to generate explanations that highlight relevant words, helping users understand the rationale behind the model's predictions. Taking inspiration from the effectiveness and interpretability of layer-wise relevance propagation~(LRP) in other short-text classification tasks~\cite{arras2017relevant,arras2017explaining,karim2021deephateexplainer}, we have adopted LRP~\cite{bach2015pixel} as the method for explaining the complex neural networks-based patent classification model.

\begin{figure}
    \centering
    \includegraphics[width=0.900\linewidth]{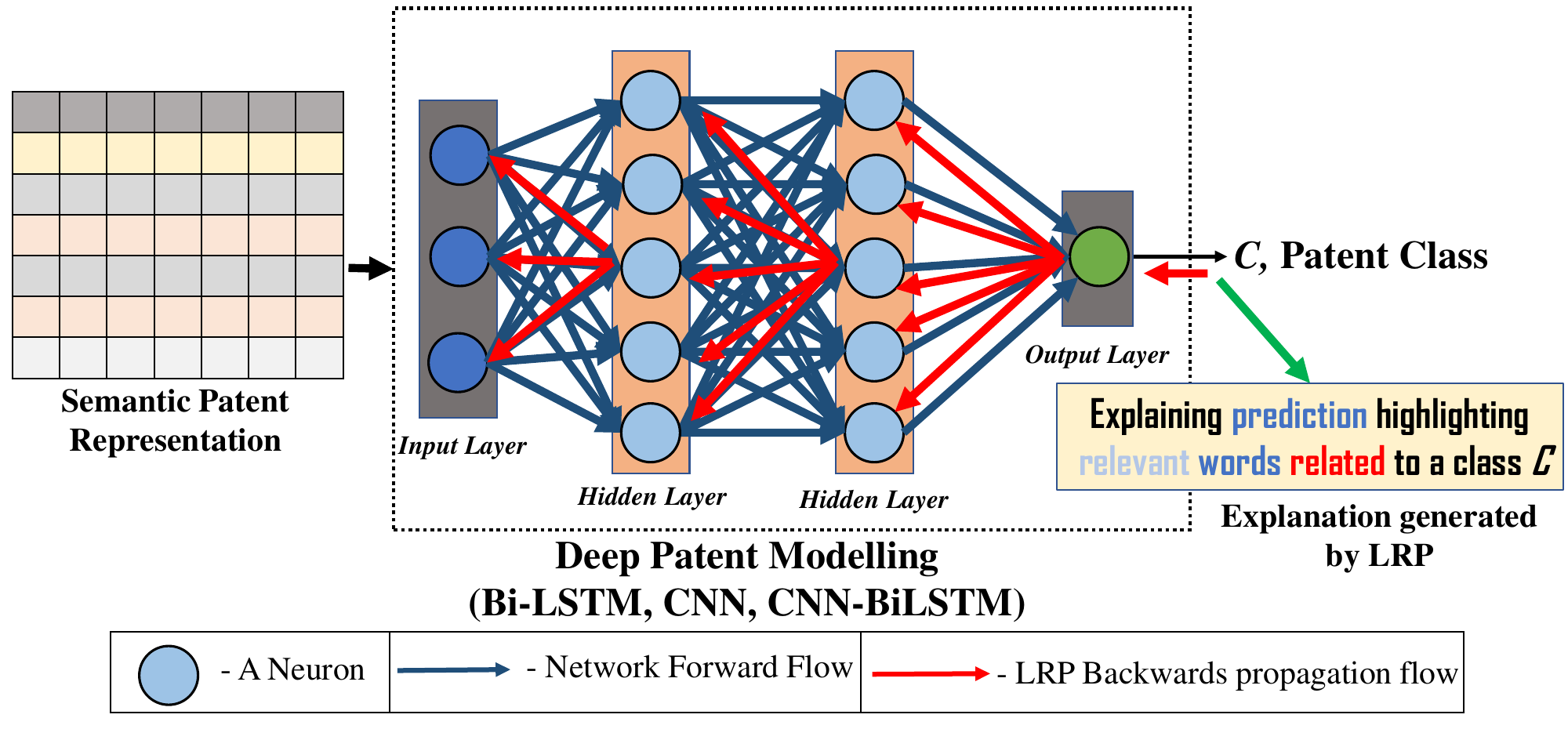}
    \caption{A conceptual overview diagram of our explainable patent classification framework.}
    \label{overview}
\end{figure}

\section{Explainable Patent Classification} \label{ourApproach}
Our proposed explainable patent classification framework consists of two major components, i) training DNN-based classification model using the semantic representation of patent text, and ii) explanation generation component leveraging layer-wise relevance propagation~(LRP). The conceptual diagram with major components is depicted in Fig~\ref{overview}. Our method first represents preprocessed patent texts semantically by high-dimensional vector leveraging pre-trained word embedding models. Then, the semantic representations for patent text are fed to train multiple DNN-based classification models including Bi-LSTM, CNN, and CNN-BiLSTM. For a particular deep patent classification model, our introduced LRP algorithm computes the relevance score towards a certain class for a given patent by redistributing the relevance score with backward propagation from the output layer to the input layer. Eventually, we get the score for patent terms that highlight the relevancy related to the predicted class of a given input patent.

\subsection{Training deep neural models}
Before training any specific DNN-based patent classification model, we employ \emph{FastText} word-embedding model to represent each word of patent text with a high-dimensional feature vector and the element of each vector carries semantic and contextual information of that word. \emph{FastText} is a character n-gram-based embedding technique. Unlike, \emph{Glove} and \emph{Word2Vec}, it can provide a word vector for out-of-vocabulary~(OOV) words. Patents' text contains less used scientific terms and some words that are higly context specific. For example, patent in the field of chemistry has a lot of reagents and chemical names, even for some new patents the reagents' names might be completely new, proposed by the inventors. Considering this intuition, we chose \emph{FastText} embedding instead of \emph{Glove} and \emph{word2vec}. We make a sequence of embedding of the words for each patent and then fed it into the deep-learning model. Our trained different neural network models includes bidirectional LSTM~(Bi-LSTM), convolutional neural networks~(CNN), CNN-BiLSTM, a combination of CNN and Bi-LSTM. 

\subsection{Explaining predictions with LRP}
Let $c$ denotes the predicted class for the input patent $p$. The LRP algorithm applies the layer-wise conservation principle to calculate the relevance score for features. The computation starts from the output layer and then redistributes the relevance weight, eventually back-propagating it to the input layers~\cite{arras2017explaining,arras2017relevant}. In other words, the relevance score is computed at each layer of the DNN model. Following a specific rule, the relevance score is attributed from lower-layer neurons to higher-layer neurons, and each immediate-layer neuron is assigned a relevance score up to the input layers, based on this rule. The flow of propagation for computing the relevance is depicted by the red arrow that goes from the output towards the input layers in Fig.~\ref{overview}. 

\if false
\begin{figure}[!htb]
    \centering
    \includegraphics[width=0.90\linewidth]{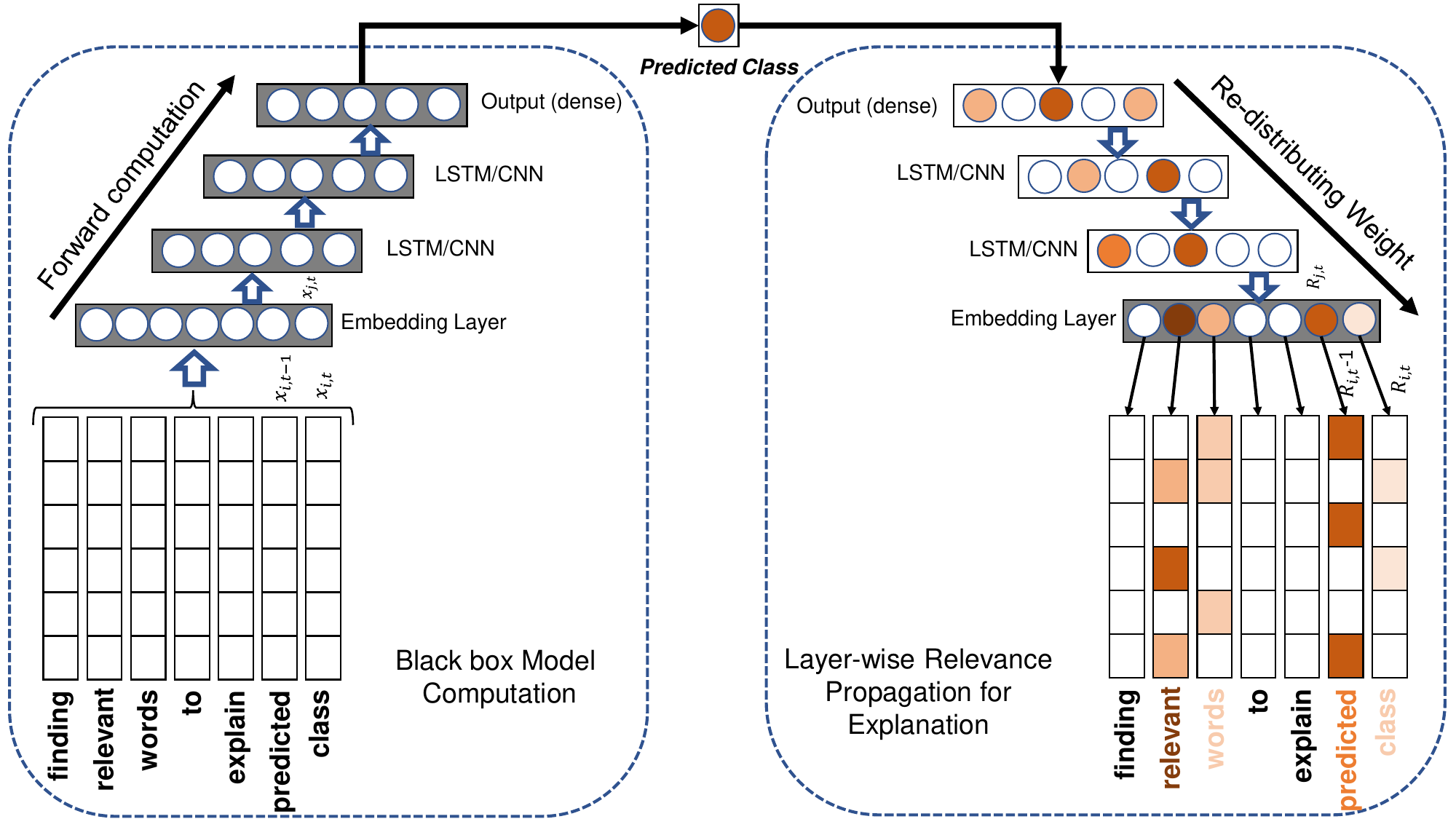}
    \caption{A conceptual overview diagram illustrating the working flow of layer-wise relevance propagation~(LRP)~(Figure created based on~\cite{arras2017relevant}).}
    \label{LRP}
\end{figure}
\fi 
The prediction score, $f_c(p)$ by our deep patent classification model, which is a scalar value corresponding to the patent class $c$. Using LRP, our aim is to identify the relevance score for each dimension $d$ of a given patent vector $p$ for the target patent class $c$. Our objective is to compute the relevance score of each input feature (i.e., words) that illustrates how positively (or negatively) contributes to classifying the patent as class $c$ (or another class). Let $z_j$ be the neuron of the upper layer and the computation of the neuron is calculated as 
\begin{equation}
    z_j = \sum_i z_i \cdot w_{ij} + b_j,
\end{equation}
where $w_{ij}$ be the weight matrix and $b_j$ denotes the bias~\cite{arras2017explaining}. Given that the relevance score for upper-layer neurons $z_j$ is $R_j$ and we move towards lower-layer neurons to distribute that relevance. In the final layer, there is only one neuron (i.e., the prediction score) and in that case, $R_j$ is the prediction score by the function $f_c(p)$. The redistribution of the relevance to the lower layers is done by following two major steps. We need to compute relevance messages to go from upper-layer to lower-layer neurons~\cite{arras2017explaining}. 

Let $i$ be the immediate lower layer and its neurons are denoted by $z_i$. Computationally, the relevance massages $R_{i \leftarrow j }$ can be computed as followings~\cite{arras2017explaining}.

\begin{equation}
   R_{i \leftarrow j } = \frac{z_i \cdot w_{ij} + \frac{\epsilon \cdot sign(z_j) + \delta \cdot b_ j} {N} }{z_j + \epsilon \cdot sign(z_j)} \cdot R_j.
\end{equation}

The total number of neurons in the layer $i$ is denoted as $N$ and $\epsilon$ is the stabilizer, a small positive real number (i.e., 0.001). By summing up all the relevance scores of the neuron in $z_i$ in layer $i$, we can obtain the relevance in layer $i$, $R_i = \sum_i R_{i\leftarrow j}$. $\delta $ can be either 0 or 1 (we use $\delta =1$)~\cite{arras2017explaining,karim2021deephateexplainer}. With the relevance messages, we can calculate the amount of relevance that circulates from one layer's neuron to the next layer's neuron. However, the computation for relevance distribution in the fully connected layers is computed as $R_{j\rightarrow k} = \frac{z_{jk}}{\sum_j z_{jk}}R_k$~\cite{arras2017relevant}. The value of the relevance score for each relevant term lies in [0,1]. The higher the score represents higher the relevancy of the terms towards the predicted class. 

\section{Experiments}\label{experiments}
This section presents the details about the datasets, experiment results, and discussion of generated explanation with LRP.

\subsection{Dataset}
\textbf{\textit{AI-Growth-Lab
patent dataset: }}We conducted experiments on a dataset containing 1.5 million patent claims annotated with patent class\footnote{Dataset: {https://huggingface.co/AI-Growth-Lab}}~\cite{bekamiri2021patentsberta}. According to the CPC patent system, the classification is hierarchical with multiple levels including section, class, subclass, and group. For example, there are 667 labels in the subclass level~\cite{bekamiri2021patentsberta}. However, for a better understanding of the generated explanations and the reasons behind a prediction for a given patent, we modeled the patent classification task with 9 general classes including \textit{Human necessities, Performing operations; transporting, Chemistry; metallurgy, Textiles; paper, Fixed constructions, Mechanical engineering; lighting; heating; weapons; blasting engines or pumps, Physics, Electricity} and \textit{General}. 
 
\vspace{2em}

\noindent \textbf{\textit{BigPatent dataset: }} BigPatent\footnote{Dataset: {https://huggingface.co/datasets/ccdv/patent-classification/tree/main}} dataset is prepared by processing 1.3 million patent texts~\cite{sharma2019bigpatent}. However, the classification dataset contains in total of 35k patent texts with 9 above-mentioned classes as labels. They provided the dataset by splitting it into training, validation, and testing set, the number of samples are 25K, 5K, and 5K, respectively. There are two different texts for each patent, one is a raw text from patent claims and another version is the human-generated abstract summarized from the patent claims.

\begin{figure}[!htb]
    \centering
    \includegraphics[width=0.90\textwidth]{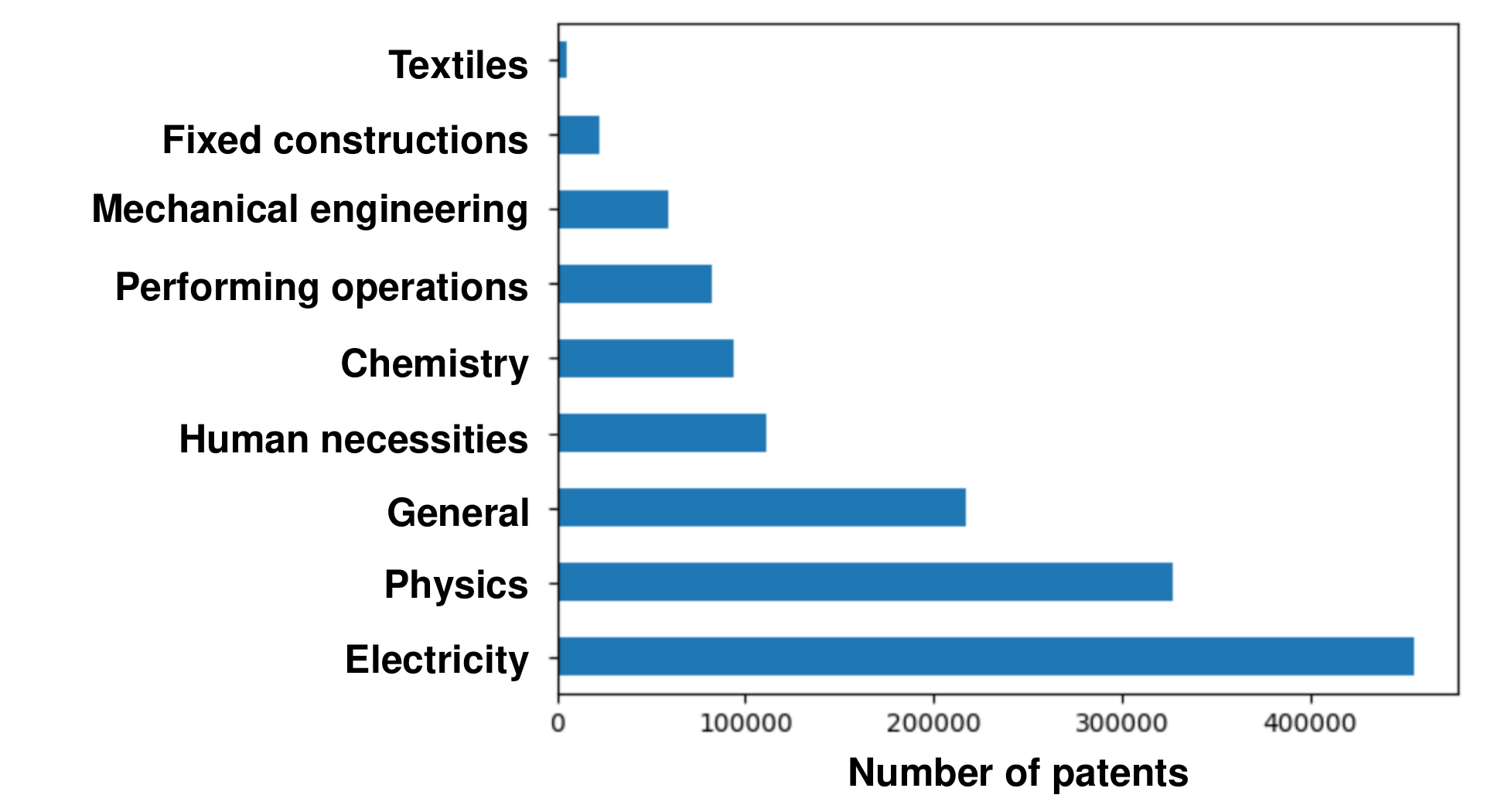}
    \caption{The distribution of the patents for different class on AI-growth-Lab data}
    \label{distribution_AIG}
\end{figure}
\begin{figure}[!htb]
    \centering
    \includegraphics[width=0.90\textwidth]{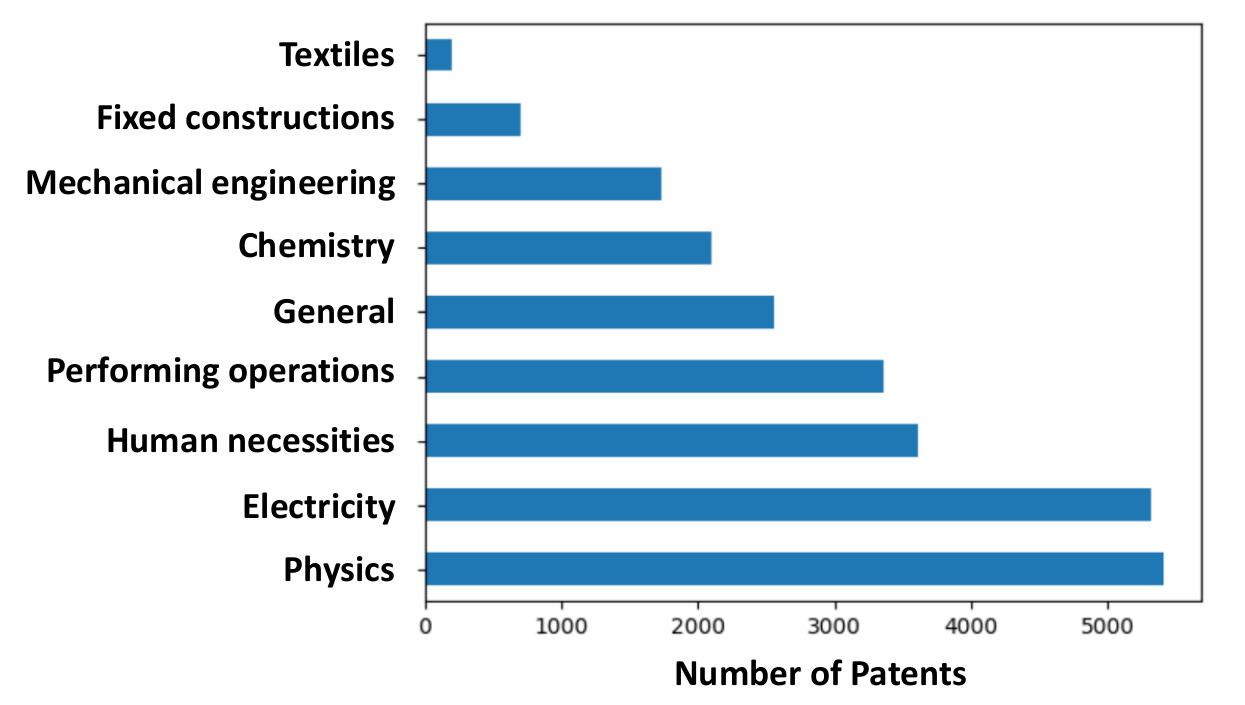}
    \caption{The distribution of the patents for different class on BigPatent data}
    \label{distribution_Big}
\end{figure}

However, the number of samples per patent class is varied widely for both both datasets, which means both are imbalanced dataset. The horizontal bar chart in Fig.~\ref{distribution_AIG} and~\ref{distribution_Big} show the level of imbalance for both datasets. This imbalance distribution of samples per class poses an additional challenge in this multi-level classification task.

\subsection{Experimental setup}
We conducted experiments using three different DNN models, namely Bi-LSTM, CNN, and CNN-BiLSTM, utilizing the \emph{FastText} pre-trained word-embedding model for text representation in the embedding layers. The Bi-LSTM model consists of a layer of Bi-LSTM with 64 units after embedding layer, followed by another Bi-LSTM layer with 32 units, and then two fully-connected layers with 64 and 9 units, respectively. We applied the rectified linear units (ReLU) activation function in the hidden dense layer, and the softmax activation function in the output layer. For the CNN model, after the embedding layer, we have a 1-dimensional convolutional layer followed by a global average pooling layer, and finally, the output layer is a fully-connected layer with 9 units. The CNN-BiLSTM model has a convolutional layer followed by a global average pooling layer, and then the Bi-LSTM part is similar to the above-mentioned Bi-LSTM model. The activation functions in the fully connected hidden and output layers are ReLU and softmax, respectively. We implemented our methods using \emph{scikit-learn} and \emph{Keras}, and represented the patent text using the \emph{FastText} pre-trained word-embedding model\footnote{https://fasttext.cc/docs/en/crawl-vectors.html}. For implementing LRP for the Bi-LSTM network, we followed the method described in~\cite{arras2017explaining}\footnote{https://github.com/ArrasL/LRP\_for\_LSTM}. For the BigPatent dataset, the training, testing, and validation sets are already split. For the AI-Growth-Lab data, the ratio for the training and testing set is 80\% and 20\%, respectively.

\begin{table}[!htb]
    \centering
    \caption{The performance of different deep patent classification models on two datasets in terms of precision, recall and f1-score. The best result is in \textbf{bold}.}
    \vspace{6px}
    \begin{tabular}{|c|c|c|c|c|}
        \hline
        \textbf{Dataset} &
        \textbf{Method} & \textbf{Precision} & \textbf{Recall} & \textbf{F1-Score} \\ \hline \hline
        \multirow{3}{*}{AI-Growth-Lab} &
        \textbf{Bi-LSTM}& \textbf{0.69}& \textbf{0.70}&\textbf{0.69}\\ \cline{2-5} 
        &\textbf{CNN}& 0.62& 0.63 & 0.62 \\ \cline{2-5}
        &\textbf{CNN-BiLSTM}& \textbf{0.69}& 0.68& \textbf{0.69} \\ \hline \hline
        \multirow{3}{*}{BigPatent} &
        \textbf{Bi-LSTM}& \textbf{0.79} & \textbf{0.78} & \textbf{0.78} \\ \cline{2-5}
        &\textbf{CNN}& 0.75 & 0.76& 0.76 \\ \cline{2-5}
        &\textbf{CNN-BiLSTM}& 0.77& 0.76 & 0.76 \\ \hline         
    \end{tabular}
    
    \label{table1}
\end{table}
\subsection{Performance analysis}
The performance of the proposed classification models was evaluated using three evaluation metrics, including Precision, Recall, and F1-Score, on two datasets, as shown in Table~\ref{table1}. The results demonstrate consistent performance across most of the deep classification models. Among them, the Bi-LSTM model exhibited better performance in terms of all evaluation metrics on both datasets. However, the performance of the other two models, CNN and CNN-BiLSTM, was also consistent and effective, though slightly lower than the Bi-LSTM model. Specifically, for the first dataset, CNN-BiLSTM performed equally well in terms of Precision (0.69) and F1-Score (0.69), while the performance of the CNN-based model was comparatively lower for the AI-Growth-Lab dataset, with a Precision of 62\%, which was 7\% lower than the best-performing Bi-LSTM model. However, for the BigPatent dataset, the CNN model exhibited considerably better performance, with a Precision of 75\%, which was only 4\% lower than the Bi-LSTM model. The performance difference between the models for the other two metrics was even lower, at 2\%.

\begin{table}[!htb]
    \centering
     \caption{Class-wise performance of Bi-LSTM model on BigPatent Dataset}
     \vspace{6px}
    \begin{tabular}{|l|c|c|c|c|}
    \hline
    \textbf{Patent Class} & \textbf{label} & \textbf{Precision} & \textbf{Recall} & \textbf{F1-score} \\ \hline \hline
     Human necessities    & 1 & 0.79   &   0.91   &   0.85 \\ \hline     
 Performing\_operations    & 2 &  0.74   &   0.66   &   0.70  \\ \hline     
             Chemistry    & 3 &  0.75   &   0.88   &   0.81  \\ \hline     
              Textiles    &  4 & 0.71   &   0.74   &   0.73  \\ \hline      
   Fixed\_constructions    &  5 & 0.65   &   0.70   &   0.67  \\ \hline     
Mechanical\_engineering    &  6 & 0.60   &   0.84   &   0.70  \\ \hline    
               Physics    &  7 & 0.75   &   0.82   &   0.78   \\ \hline    
           Electricity    &  8 & 0.78   &   0.86   &   0.82   \\ \hline   
               General    &  9 & 0.71   &   0.46   &   0.41   \\ \hline 
    \end{tabular}
    \label{class_wise}
\end{table}

The performance of all DNN-based classifiers on the BigPatent dataset is significantly superior compared to the first dataset. This may be attributed to the fact that the BigPatent dataset includes finely-grained abstracts of patents which are generated by human assessors, taking into consideration the patent texts. As a result, the semantic representation of the fine-tuned text in the BigPatent dataset is enriched compared to the raw patent claims in other dataset. We present the performance of Bi-LSTM model by showcasing the class-wise performance on the BigPatent dataset. Table~\ref{class_wise} displays the performance across nine different patent classes. The Bi-LSTM model demonstrates favorable and consistent performance across most patent classes, with the exception of the \textit{general} category. It is hypothesized that the patents in the \emph{``general''} category may contain more commonly used terms compared to patents in other area-specific categories. Consequently, the captured semantic information may not be sufficient, potentially resulting in lower performance in terms of recall and F1-Score for the \emph{``general''} class compared to other classes.

\begin{table}[!htb]
    \centering
    \caption{Performance comparison with related works}
    \begin{tabular}{|c|c|c|c|}
    \hline
    \textbf{Method} & \textbf{Precision} & \textbf{Recall} & \textbf{F1-Score} \\ \hline \hline
    Out Method& 0.79 & \textbf{0.78} & \textbf{0.78} \\ \hline
     Roudsari et al.~\cite{haghighian2022patentnet} (Bi-LSTM)& 0.7825 &0.6421 &0.6842 \\  \hline
    Roudsari et al.~\cite{haghighian2022patentnet} (CNN-BiLSTM) & 0.7930 & 0.6513 & 0.6938\\ \hline
    Shaobo et al.~\cite{li2018deeppatent} (DeepPatent) &\textbf{0.7977} & 0.6552 & 0.6979\\ \hline
         
    \end{tabular}
    \label{comparison_with_related_work}
\end{table}

We compared the performance of our models with similar models that used \emph{FastText} embedding for patent text representation. Compared to existing works by Roudsari et al.~\cite{haghighian2022patentnet} and Shaobo et al.~\cite{li2018deeppatent}, the performance of our trained models is effective. Roudsari et al. also trained similar models with semantic text representation with a pre-trained \emph{FastText} word-embedding model. They also develop similar DNN models including Bi-LSTM and CNN-BiLSTM.  Shaobo et al.~\cite{li2018deeppatent} introduced CNN-based deep patent modelling employing \emph{FastText} word-embedding model. The performance of our methods on BigPatent data is higher than their models for all evaluation metrics except Precision. The comparison shows the effectiveness of our methods in classifying patents.

\subsection{Generated explanation for prediction}
\begin{figure}[!htb]
    \centering
    \includegraphics[width = 0.90\textwidth]{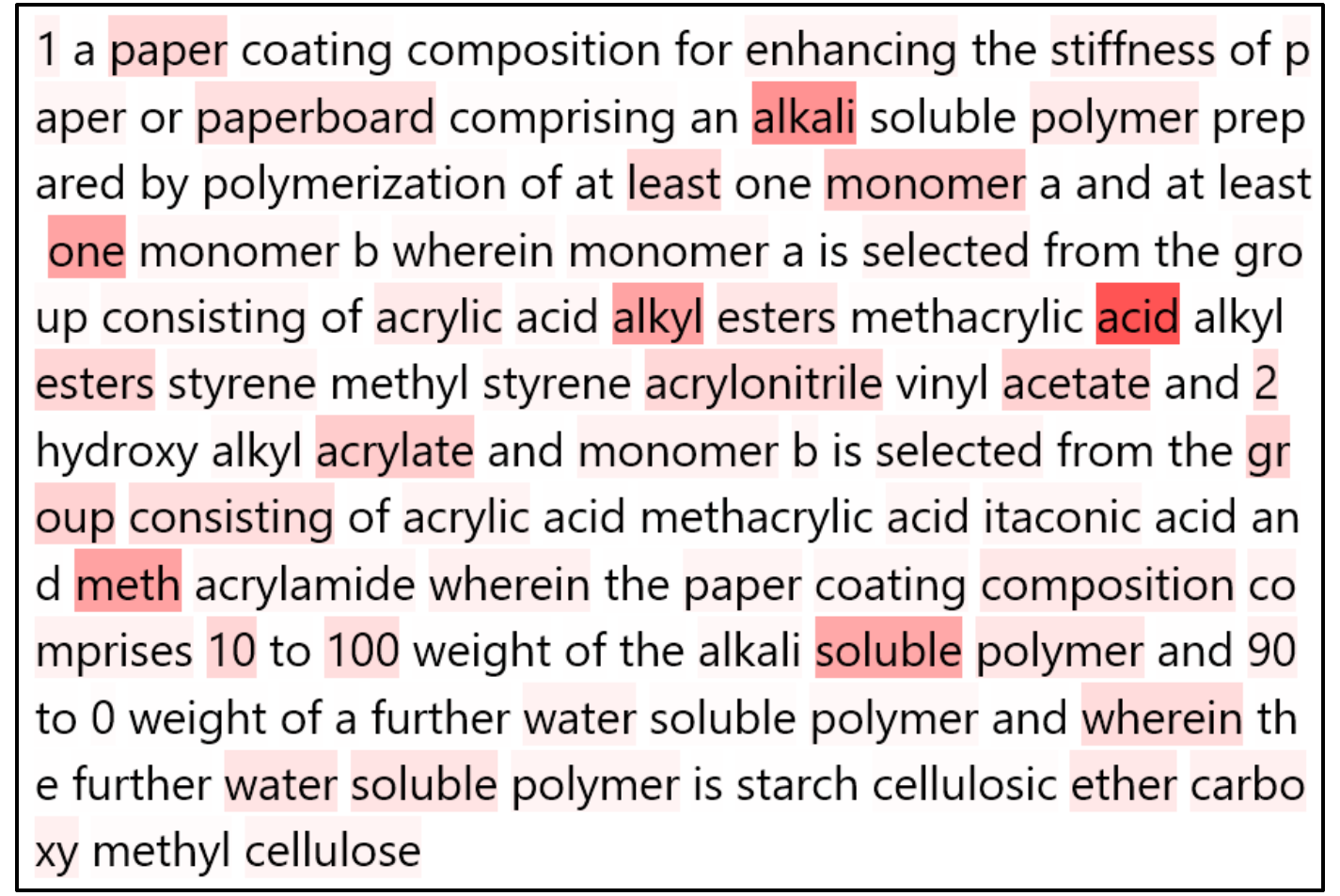}
    \caption{An example explanation for a patent classified as \emph{Chemistry} patents highlighting relevant words. The higher the intensity of the color, the better the relevancy of the words contributing to the prediction.}
    \label{explan1}
\end{figure}

We attempted to unbox the black-box nature of the deep patent classification model by adopting a layer-wise relevance propagation technique to compute the relevance score for each term by back-propagating the prediction score from the output layer to input layers. To represent the explanation per predicted class for a given patent text, we highlighted the related words that contributed to the classifier's prediction. As an example explanation, a patent is classified as \emph{Chemistry}, and the related words that contributed to the prediction are highlighted in red color in Fig~\ref{explan1}. The figure shows the explanation highlighting relevant words for the patent that classified as \textit{chemistry}. The intensity of the color represents the contributions of a particular word. The higher the intensity of the color~(red), the better the relevancy the word is. 
\begin{figure}[!htb]
    \centering
    \includegraphics[width = 0.90\textwidth]{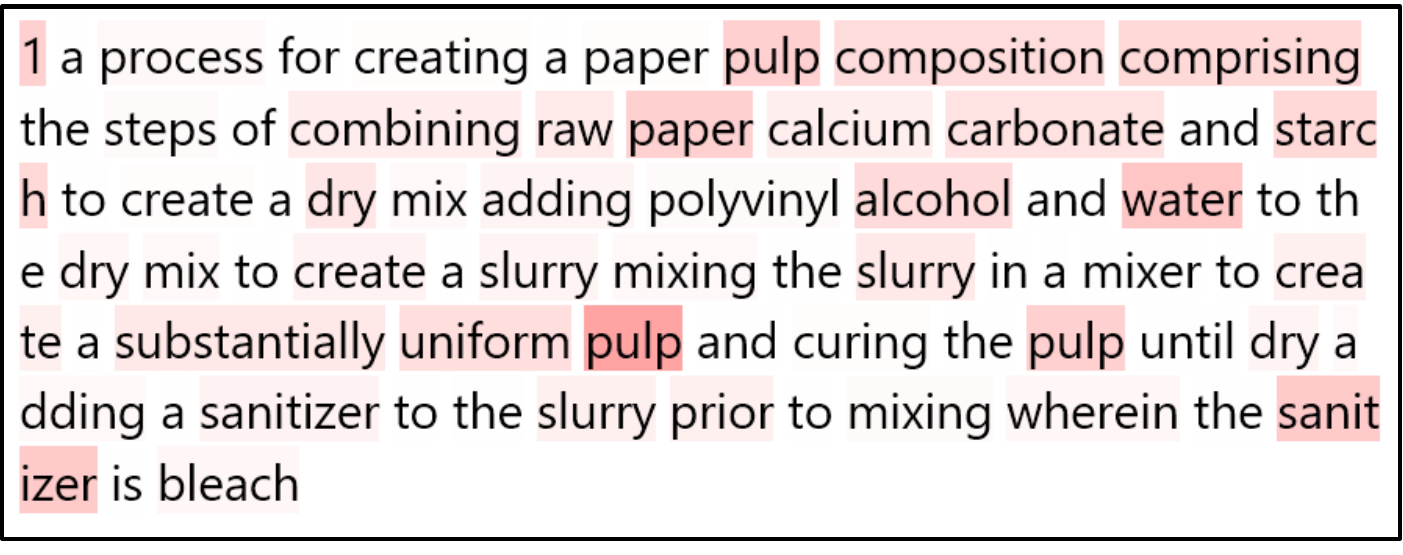}
    \caption{An example explanation for a patent classified as \emph{Chemistry} patents highlighting relevant words. The higher the intensity of the color, the better the relevancy of the words contributing to the prediction.}
    \label{explan2}
\end{figure}
We can see that from the figure, the most relevant words include, \emph{alkali, alkyl, monomer, acid, acrylate, acrylonitrile, acetate, polymer, ether}. We can observed that the highlighted words are completely related to terms used in organic chemistry and the explanation makes sense why this patent has been classified as a chemical patent. The next relevant list of words is \textit{soluble, water, stiffness, enhanching, etc}. These words are directly related to chemistry except \emph{stiffness} and \emph{enhancing}. Since \emph{enhancing} the \emph{stiffness} of the paper or paperboard is the objective of this patent, these words are selected as relevant. 
\if false
\begin{figure}[!htb]
    \centering
    \includegraphics[width = 0.90\textwidth]{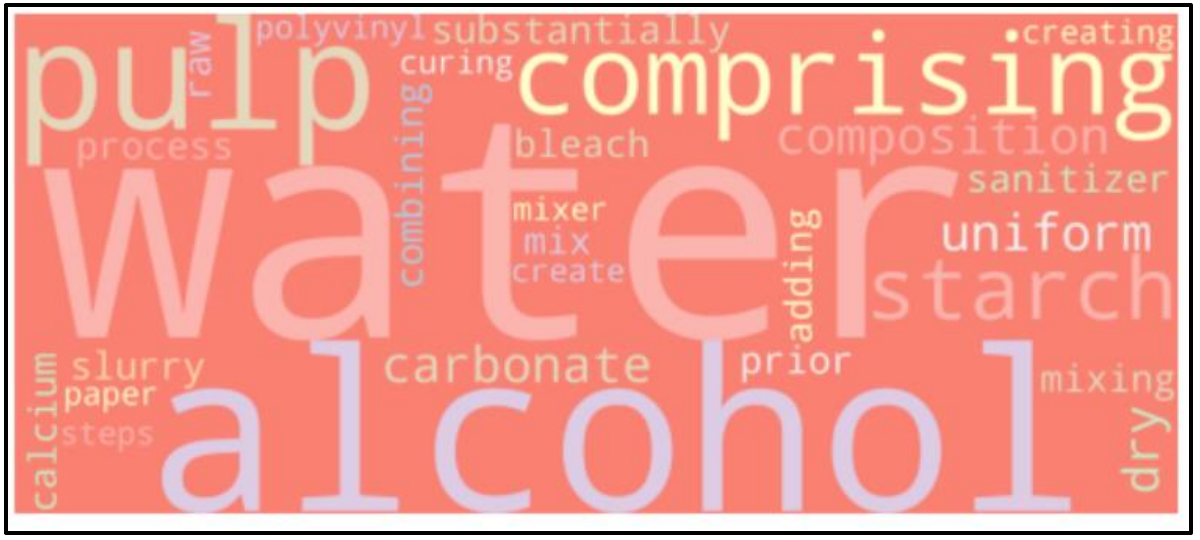}
    \caption{An example explanation for a patent classified as \emph{Chemistry} patent highlighting relevant words in word cloud. The larger the font of the word, the better the relevancy of the words contributing to the prediction.}
    \label{wordcloud1}
\end{figure}
\fi

\begin{figure}[!htb]
    \centering
    \includegraphics[width = 0.90\textwidth]{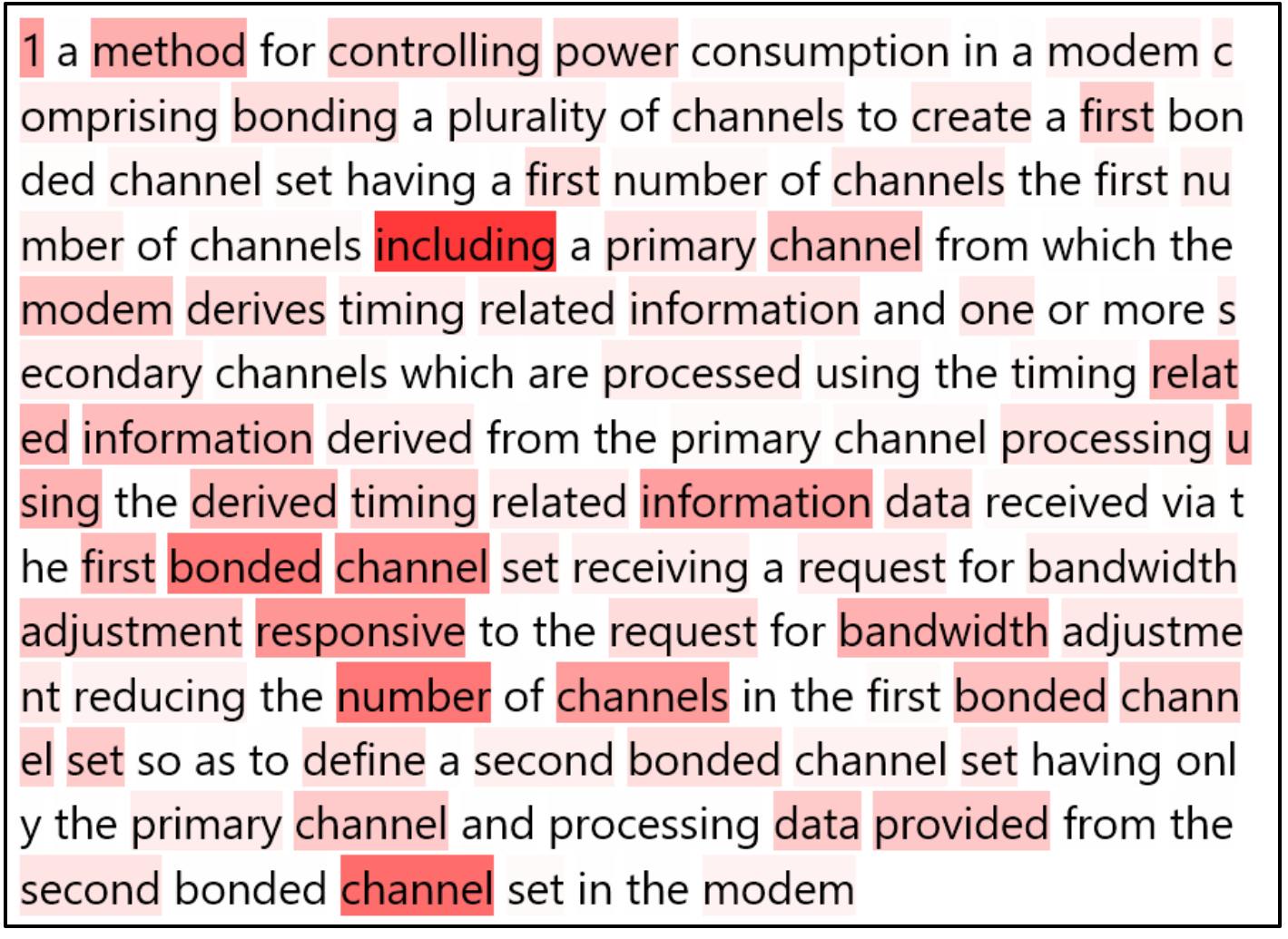}
    \caption{An example explanation for a patent classified as \textit{Electricity} patents highlighting relevant words. The higher the intensity of the color, the better the relevancy of the words contributing to the prediction.}
    \label{explan3}
\end{figure}

For another example patent in the field of \textit{Electricity}, Fig.~\ref{explan3} illustrates the explanations highlighting relevant words that contributed to the classifier to decide that the patent is from \emph{electricity} field. The most relevant words, in this case, include \emph{power, channel, modem, device, bonded, bandwidth, data,} etc. We can see that all identified related words are used in \emph{electricity} literature. The word \emph{device} is used for common use in some other fields also, but this word also can be used to mention any electrical instrument in electricity-related explanation. 
However, there are some words selected as relevant for both examples which are not relevant to the specific fields but can be used in literature for any field. One plausible reason is that those also might carry considerable importance in describing the any scientific object (i.e., explaining chemical reaction) and capture good contextual and semantic information in \emph{FastText} embedding. 
\if false
\begin{figure}[!htb]
    \centering
    \includegraphics[width = 0.90\textwidth]{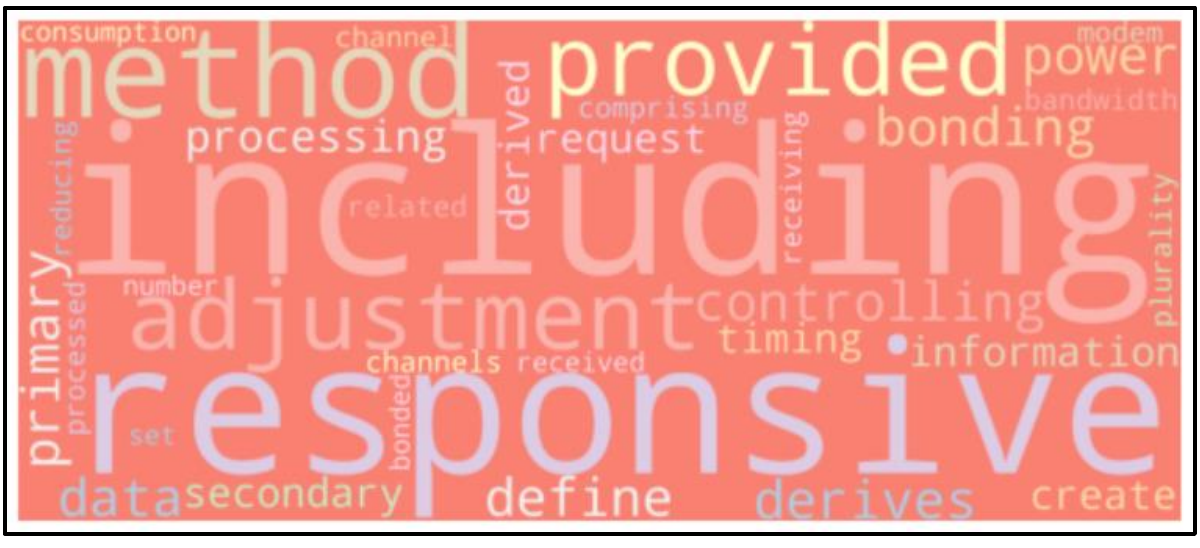}
    \caption{An example explanation for a patent classified as \emph{Electricity} patent highlighting relevant words in word cloud. The larger the font of the word, the better the relevancy of the words contributing to the prediction.}
    \label{wordcloud2}
\end{figure}
\fi 

\subsection{Limitations}
Our model can explain the prediction for multi-label classification. Since the patents are classified in different levels and the patent classification system has a huge set of classes to classify in different levels, it should be explainable for multi-level classification also. This will be more challenging to explain the prediction for different subgroups-level classes. Another limitation is that our utilized pre-trained word-embedding model is not trained on the patent corpus. The local word-embedding model trained with patent corpus might capture better contextual and semantic information for scientific terms and jargon. Hence, the performance might be better than the current approach.

\section{Conclusion and Future Direction}\label{conclusion}
This paper aimed at explaining the predictions from DNN-based patent classification models with layer-wise relevance propagation technique to identify the relevance of different words in the patent texts for a certain predicted class. Layer-wise relevance propagation technique can capture context-specific explanatory and relevant words to explain the predictions behind certain predicted classes. The experimental results demonstrated the effectiveness of classifying patent documents with promising performance compared to existing works. We observed that the explanations generated by the LRP technique make it  easier to understand why a certain patent is classified as a specific patent class. Most of the captured words have high relevancy with the patent domain, even though a few words marked as related are not that relevant (which, however, should also provide useful information to human expert in assessing the predictions). Even though our approach would still need to be evaluation with users, we can observe that the explanations are helpful to understand the question why a certain patent was classified into a specific class, and to assess the results of deep-learning-based complex artificial intelligence-enabled models. 

Since patents have a lot of scientific and uncommon words and phrases (i.e., jargon) that are not often used in other texts, we plan to train a local word-embedding model with patent texts to have better representation in our future work. It would be interesting to apply a transformer-based approach for the same purpose. The explanations for sub-group level prediction and capturing the sub-group context will be even more explanatory. However, the generated explanations will need to be evaluated by human experts in the patent industry. Therefore, we plan to have a user-centric evaluation for the generated explanations and elicit more human-centric requirements to be addressed in the future for better adoption real-word applications. 

\section*{Acknowledgment}
This project has received funding from the European Union's Horizon 2020 research and innovation programme under the Marie Skłodowska-Curie grant agreement No 955422.

\bibliographystyle{unsrtnat}
\bibliography{references}  

\end{document}